\documentclass[english]{article}

\usepackage[T1]{fontenc}
\usepackage[latin9]{inputenc}
\usepackage{amsmath}
\usepackage{graphicx}

\makeatletter

\providecommand{\tabularnewline}{\\}

\numberwithin{equation}{section}
\numberwithin{figure}{section}
\newcommand{\lyxaddress}[1]{
\par {\raggedright #1
\vspace{1.4em}
\noindent\par}
}

\makeatother

\usepackage{babel}
\begin{document}

\title{Artificial Intelligence Approaches To UCAV Autonomy}

\author{Amir Husain\thanks{SparkCognition Inc.}, Bruce Porter\thanks{Department of Computer Science, University of Texas at Austin.}}
\maketitle

\lyxaddress{\begin{center}
amir@sparkcognition.com, porter@cs.utexas.edu
\par\end{center}}

\section{Abstract}

This paper covers a number of approaches that leverage Artificial
Intelligence algorithms and techniques to aid Unmanned Combat Aerial
Vehicle (UCAV) autonomy. An analysis of current approaches to autonomous
control is provided followed by an exploration of how these techniques
can be extended and enriched with AI techniques including Artificial
Neural Networks (ANN), Ensembling and Reinforcement Learning (RL)
to evolve control strategies for UCAVs.

\section{Introduction}

Current UAVs have limited autonomous capabilities that mainly comprise
GPS waypoint following, and a few control functions such as maintenance
of stability in the face of environmental factors such as wind. More
recently some autonomous capabilities such as the ability for a fixed
wing UCAV to land on the deck of a carrier have also been demonstrated
\cite{key-2}. These capabilities represent just the tip of the spear
in terms of what is possible and, given both the commercial and military
applications and interest, what will undoubtedly be developed in the
near future. In particular, flexibility in responses that can mimic
the unpredictability of human responses is one way in which autonomous
systems of the future will differentiate themselves from rules-based
control systems. Human-style unpredictability in action selection
opens the door to finding solutions that may not have been imagined
at the time the system was programmed. Additionally, this type of
unpredictability in combat systems can create difficulties for adversary
systems designed to act as a counter.

The capability to compute sequences of actions that do not correspond
to any pre-programmed input - in other words, the ability to evolve
new responses - will be another area of future differentiation. There
are many other such enhancements that will be enabled via autonomous
systems powered by Artificial Intelligence. In the following sections
we will outline some of the advanced capabilities that can be engineered,
and design and engineering approaches for these capabilities. 

\section{Existing Control Systems}

Some degree of autonomy in flight control has existed for over a hundred
years, with autopilot inventor, Lawrence Sperry's demonstration in
1913 \cite{key-6} of a control system that tied the heading and attitude
indicators to a control system that hydraulically operated elevators
and rudders. A fully autonomous Atlantic crossing was achieved as
early as 1947 in a USAF C-54 aircraft \cite{key-7}. However, much
of the early work in automating control systems were mechanical implementations
of rule-based systems drawing upon cybernetics and control theory.
They demonstrated that with such techniques it was possible to automate
a basic mission, including takeoff and landing. 

Since the 1947 demonstration, considerable effort has been invested
in developing autonomous flight capabilities for commercial and military
aircraft. Modern flight control or autopilot systems that govern landings
are segmented in five categories from CAT-I to CAT-IIIc \cite{key-12},
with capabilities varying based on forward visibility and decision
height. Many of these systems use rule-based, or fuzzy-rule based
control, incorporating sensor-fusion techniques such as Kalman filters
\cite{key-8}. They are capable of following a planned route and adjusting
for environmental factors such as cross-winds, turbulence and so on. 

The increased popularity of commercial drones, and the heightened
utilization of military drone aircraft has, in parallel, created a
new class of autonomous capabilities. From Open Source initiatives
such as the Ardupilot\cite{key-9}flight control software for low-cost
drones, to higher levels of autonomy in military drones. Software
such as the Ardupilot, for example, uses a combination of GPS positioning,
additional sensors to gauge velocity and position, combined with basic
flight control rules to autonomously navigate to a sequence of waypoints.
Many of these map-input based waypoint following capabilities are
also implemented in military surveillance and combat drones.

Another area of control innovation comes from swarm theory and related
control algorithms. At the simplest level, these algorithms seek inspiration
from the behavior of biological systems such as ant colonies or flocks
of birds. They are collaboration algorithms that enable each individual
system in the swarm to compute its future actions based on its own
measurements, but also those of its neighbors. While basic swarm algorithms
\cite{key-11} are effective in providing coverage over an area, and
automatically repositioning all nodes when one is lost to maintain
coverage, they do not provide much guidance on how to divide mission
responsibilities and burdens, and to effectively delegate them to
individual nodes. The concept of a ``swarm'' as found in biology
will have to evolve into something entirely different - perhaps somewhat
similar to a pack hunt - but even that analogy would only be marginal
- in order for it to be an effective and useful system particularly
in a military context. Some of the reasons why we propose this conclusion
regarding the inadequacy of existing swarm algorithms is that most
biologically inspired algorithms, such as Particle Swarm Optimization
(PSO) \cite{key-13} or Artificial Bee Colony Algorithm (ABC) \cite{key-14},
are search or optimization techniques that do not account for the
role of an individual particle (or node) in the swarm. For example,
PSO proposes the same meta-heuristic for computing positional updates
for all points and does not incorporate a differential update mechanism
based on the role of a particle. In a subsequent publication, we intend
to propose a ``Pack Hunt Optimization'' (PHO) algorithm that we
believe addresses the shortcomings of the existing swarm algorithms
we have cited, and holds relevance to UCAV control applications.

The state of current control systems can be summed up as follows:
\begin{itemize}
\item Effective at basic navigation and path following
\item Many existing techniques to fuse sensor data for accurate position
identification
\item Able to automatically take off and land if runways are properly instrumented
\item Actions beyond flight control (such as weapons engagement) are presently
manual
\item Missions are pre-defined
\item Swarm algorithms can provide additional value for relative positioning
of multiple assets and distributed sensing
\end{itemize}

\section{Advanced Autonomous Capabilities}

The purpose of this section is to outline a few areas of potential
advancement that can be expected of autonomous systems of the future.
This list is neither exhaustive nor complete with regards to the author's
current conception of all such advanced capabilities. It is a subset
of possible functions that is listed to illuminate the broad contours
of what is possible in terms of applications of Artificial Intelligence
to UCAV autonomy. Some features include:
\begin{enumerate}
\item Knowledge \& Assessment Updates

\begin{enumerate}
\item Identification of potential threats outside pre-programmed mission
briefs
\item Autonomous exploration and assessment of identified targets that autonomous
control deems to be high priority
\item Enhancement and update to intelligence supplied as part of the mission
brief and plan, based on actual observation
\end{enumerate}
\item Autonomous Navigation and Swarm Coordination

\begin{enumerate}
\item Ability to adjust to environmental conditions that cause system or
any linked swarm systems to deviate from mission plan expectations
\item Ability to adjust to loss of a Swarm asset, not just in terms of re-positioning,
but including potential re-tasking (i.e. assumption of a new role
on the part of an individual asset)
\end{enumerate}
\item Autonomous Evasion 

\begin{enumerate}
\item Automated update to mission plan based on sensor detection of probable
manned aerial intercept
\item Automated update to mission plan based on detection of unexpected
sensor presence 
\item Autonomous evasion in the event of a RWR (Radar Warning Receiver)
activation or MAW (Missile Approach Warning) system activation
\end{enumerate}
\item Autonomous Targeting

\begin{enumerate}
\item Autonomous addition to target lists based on computer vision or alternate
sensor based identification of threats to mission (including surface
to air threats)
\item Autonomous addition to target lists in the event that primary targets
have already been neutralized
\item Autonomous deletion of a target from target lists in the event it
has been already neutralized, is found to violate a ``hard'' policy
constraint or is low priority and its neutralization harms the overall
achievement or success of the mission
\end{enumerate}
\end{enumerate}

\section{The Need for a New Approach}

In the preceding sections we explored the current state of autonomous
systems and the rules-based approach that is often employed to develop
these systems. Further, we also considered a number of advanced capabilities
that would be desirable in future autonomous control systems. A fundamental
challenge in developing these future capabilities is that the range
of scenarios an autonomous system would have to contend with in order
to effectively execute the required maneuvers are enormous. Tackling
such a large range of possibilities with a rules-based system will
be impractical not only because of the combinatorial explosion of
possibilities that would require individual rules, but also because
human designers of such a system may simply not be able to conceive
every imaginable scenario the autonomous system could find itself
in.

Another challenge is that rules-based systems are hard coded to measure
certain criteria, or sensor values, and then act based on this pre-specified
criteria. This hard coding means that each rule is tied to a specific
set of sensors. If additional sensors are added to a system, or existing
sensors are upgraded, a large number of rules would have to be re-written,
creating an obvious cost and effort burden.

What we have described above is far from an exhaustive list of limitations
in current autonomous systems, but we believe they are sufficient
to motivate the need for a new architecture for autonomy. A future
system that moves beyond rules-based systems, incorporates learning
capabilities so that actions can be learned rather than hard coded,
and can adapt to new information from new or better sensors, will
represent a substantial advance. In the sections that follow, we define
the contours of just such a system.

\section{An Architecture for Advanced Autonomy}

The fundamental architecture we propose in this paper is based on
multiple independent control systems connected to an action optimizer
neural network. Each of the multiple independent control systems can
be neural networks or non-ANN rule based control systems that output
a suggested vector of actions or control activations. The action optimizer
ANN gates and weighs the inputs supplied by each independent control
system.

\begin{center}
\includegraphics[scale=0.6]{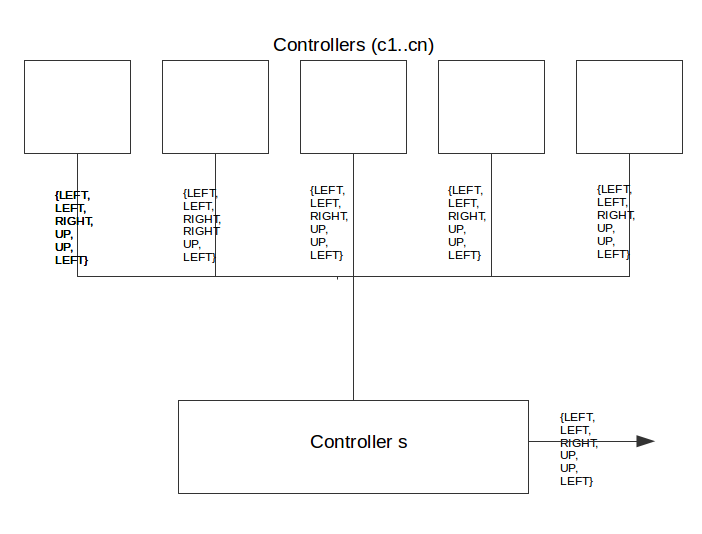}
\par\end{center}

Let $c_{k}$be an independent control system, and $s$ be an action
optimizer neural network to which $c_{1..n}$control networks are
connected. Additionally, let the set $E$ contain a collection $e_{1..m}$of
environmental inputs that are supplied to $s$. Then, we denote the
specific configuration of all environmental inputs at time $t$ by$E^{t}$and
the output of $s$ under these environmental inputs and based on the
inputs of all independent control networks, as follows:

\[
s(E^{t},C^{t})=A^{t}
\]

The goal of our system is to optimize the selection of action sequences
$A^{t..t+k}$such that this sequences maximizes the performance of
the system being controlled.

It is important to understand what we mean by, ``performance'' here.
We define performance as a variable that is the output of a utility
function $U$ such that this output is high when the weighted achievement
of all mission parameters is large, and low when the weighted achievement
of mission parameters is small. In other words, we are attempting
to locally maximize $U$ at least locally:
\[
\frac{dU}{dx}=0
\]

and:

\[
\frac{d^{2}U}{dx^{2}}<0
\]

The question obviously arises, how do we build the function $s?$
Conventionally, control functions have been built in various ways,
for example as fuzzy rule based systems \cite{key-4}. However, we
propose to implement the control function $s$ as an Artificial Neural
Network (ANN). As the application at hand will benefit from some knowledge
of past actions, we specifically propose to implement the network
as a Recurrent Neural Network (RNN).

\section{Evolving Mission Specific Controllers}

The actual training and evolution of the RNN represented by $s$ is
not the subject of this paper and will be documented in a subsequent
publication. In summary, this can be done in a manner that combines
real world and simulator environments. However, in a more detailed
future exploration we intend to cover questions such as whether individual
control networks, $c_{1..n},$ can be trained independently and how
a training set that reflects key the wide range of scenarios the UCAV
might experience would be compiled. For the purpose of the present
discussion, our basic approach is to use Reinforcement Learning (RL)
techniques \cite{key-5} to train the RNN in a simulated environment
until a basic level of competence has been achieved, and to then allow
the evolved network to control a real craft. Collected data from the
actual flight is reconciled with the simulated environment and the
process is repeated until an acceptable level of capability is demonstrated
by $s.$ This reconciliation would benefit from applications of Transfer
Learning \cite{key-15}.

One of the benefits of this approach is that the simulated environment
can introduce environmental constraints that $s$ must respond to
appropriately. For example, these can be navigation constraints such
as avoiding certain pre-identified objects on a map. Work has already
been done to use search algorithms such as A{*} to find viable paths
around objects to be avoided\cite{key-3}and this type of constraint
can be implemented by one of the independent control networks ($c_{k},$as
presented in the previous section). Other examples of existing work
that could be leveraged in the form of an independent control network
include collaborative mapping algorithms for multiple autonomous vehicles
\cite{key-16}. Of course, other constraints and optimizations would
be represented by other ensembled control networks, forcing $s$ to
weight them and choose from them carefully, in a way that maximizes
$U.$

Thus, the controller can be evolved to optimize operation in different
types of environments, and under different constraints. It may then
become possible to simply ``upload'' the optimal controller for
a particular environment, or a particular mission type, into the same
craft and achieve mission-specific optimal performance.

\section{Semantic Interpretation of Sensor Data}

Sensor data in autonomous systems does not have to remain limited
to environmental measurements or flight sensor readings. It can include
a variety of image feeds from forward, rear or down-facing cameras.
Additionally, radar data and Forward Looking Infra Red (FLIR) sensor
data is also a possibility. In order to utilize all this diverse data
to make decisions and even deviate in small but important ways from
the original mission plans, all of this data has to be interpreted
and semantically modeled. In other words, its meaning and relevance
to the mission and its role in governing future action has to be established. 

For the purpose of understanding how such data can be interpreted
and what its impact on decisions can be, we classify sensors and data
sources into the following categories:
\begin{enumerate}
\item Internal Sensors

\begin{enumerate}
\item System Health (e.g. Engine Vibration, Various Temperature and internal
system Pressure)
\item System Performance (e.g. Velocity, Stress) 
\end{enumerate}
\item External Sensors

\begin{enumerate}
\item Navigational Aides (e.g. Level, Wind speed, Inertial navigation gyroscopic
sensors)
\item Environmental Mapping (e.g. Camera, Radar, LIDAR, FLIR, RWR, MAWS)
\end{enumerate}
\end{enumerate}
In an example table below, we show the types of impact that information
received from these sensors can potentially have on mission plans
and vehicle navigation.

\begin{tabular}{|c|c|c|c|c|}
\hline 
Actions & Health & Perf. & Nav. & Envir. Mapping\tabularnewline
\hline 
\hline 
Terminate Mission & X & X &  & X\tabularnewline
\hline 
Update Mission Achievement &  &  &  & X\tabularnewline
\hline 
Add New Target &  &  &  & X\tabularnewline
\hline 
De-prioritize Target & X &  &  & X\tabularnewline
\hline 
Change Course & X &  & X & X\tabularnewline
\hline 
Add Obstacle (Constrain Path) &  &  &  & X\tabularnewline
\hline 
Engage Weapon System &  &  &  & X\tabularnewline
\hline 
Evasive Maneuvers &  &  &  & X\tabularnewline
\hline 
Engage Countermeasures &  &  &  & X\tabularnewline
\hline 
\end{tabular}

In order to support the types of advanced autonomy outlined in Section
4 of this paper, many of the actions highlighted in the table above
will likely need to be combined based on sensor input to form a chain
of actions that update the internal state and maps used by the autonomous
asset. Sensor data may be an input required by any controller $c_{k}$or
by the controller $s.$Thus, a sensor bus connects all sensors to
all controllers. 

For many sensor types, instead of the sensor providing a raw output,
we transform the output to reflect semantic constructs. For example,
instead of a raw radar signal input, we may transform the signal into
a data structure that reflects the position, speed, heading, type
and classification of each detected object. This transformation of
raw sensor data into semantic outputs that use a common data representation
for each class of sensor enables replaceability of underlying components
so that the same controllers can work effectively even when sensors
are replaced or upgraded.

The semantic output of individual sensor systems can be used by controllers,
and is also stored in a Cognitive Corpus, which is a database that
can store mission information, current status, maps, objectives, past
performance data and not-to-violate parameters for action that are
used to gate the final output of the controller $s.$

\section{Knowledge Representation For Advanced Autonomy}
\begin{center}
\includegraphics[scale=0.4]{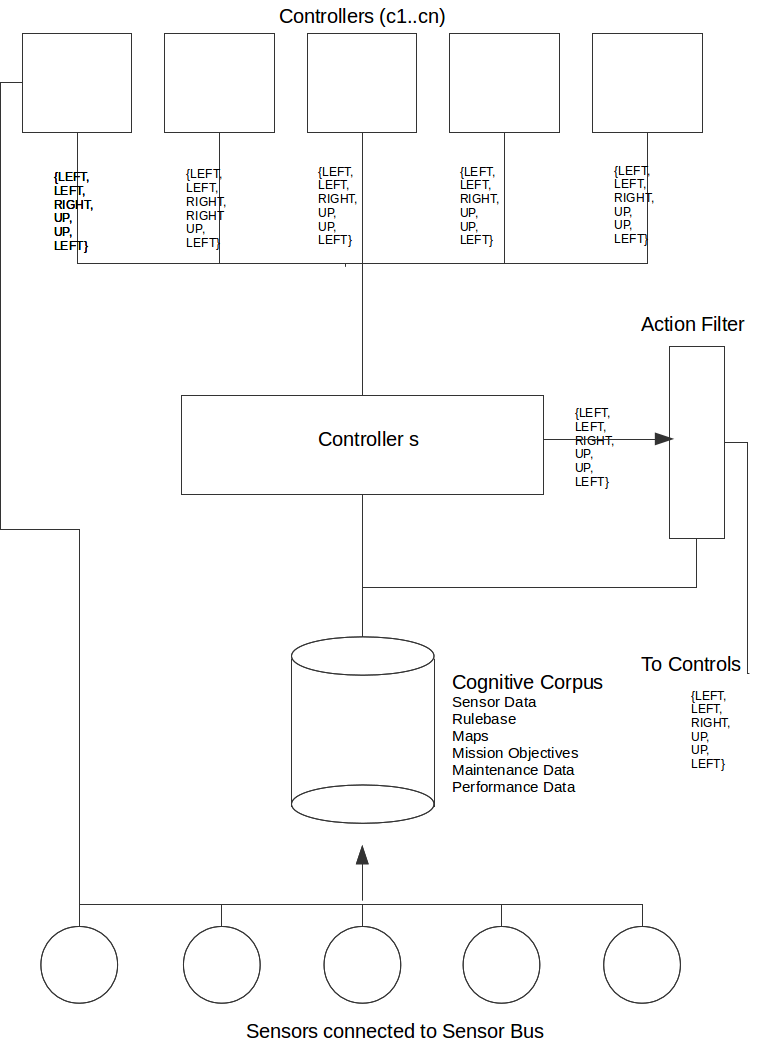}
\par\end{center}

As the more complete diagram of the proposed Autonomy Architecture
illustrates, the controller $s$ receives input from a set of controllers
$c_{1..n}$and is also connected to the sensor bus and the Cognitive
Corpus. A state map stored in the Cognitive Corpus reflects the full
environmental picture available to the autonomous asset. For example,
it includes an estimate of the asset's own position, the positions
of allied assets, the positions of enemy assets, marked mission targets,
paths indicating preferred trajectories at the time of mission planning,
territory and locations over which to avoid flight and other pertinent
data that can assist with route planning, objective fulfillment and
obstacle avoidance.

This state map forms another important input to the controller $s$
as it chooses the most optimal sequence of actions. The image below
shows a visual representation of what the state map might track. Here,
it shows the location of multiple allied assets, for example systems
that might be part of a swarm with the UCAV that is maintaining this
map. There is also a hostile entity identified with additional information
regarding its speed and heading. Locations on the ground indicate
sites to be avoided. Sensor information carried in the set (or vector)
$E$ result in updates to the state of each object in this map. Note
that the state map is maintained by each autonomous asset and while
the underlying information used to update it may be shared with, or
received from other systems, each autonomous asset acts based on its
own internal representation, or copy, of the state map.

While the details of an implementation are beyond the scope of this
paper, we propose that the information exchange between autonomous
systems occur using a blockchain protocol \cite{key-10}. Benefits
of this approach include the fact that in the event communication
is interrupted and updates are missed, information can be reconstructed
with guarantees regarding accuracy and order. Further, the use of
a blockchain store ensures that a single or few malicious participants
cannot impact the veracity of the information contained therein. 
\begin{center}
\includegraphics[scale=0.6]{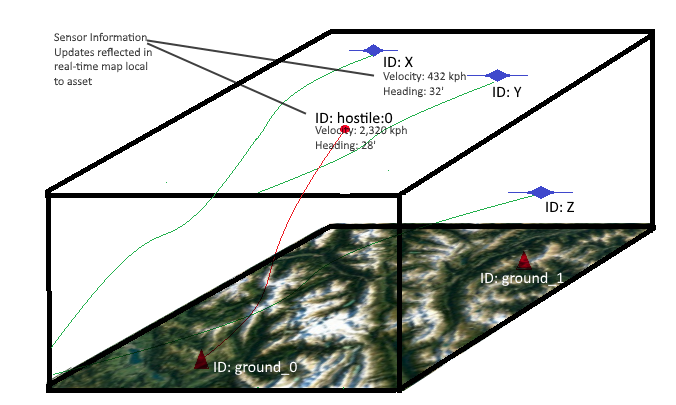}
\par\end{center}

While the figure shows a graphic representation of the map, it is
possible to represent such a map as a vector or matrix. By so doing,
it can readily be supplied to the controller $s$ as an input.

\section{Conclusion}

Sophisticated autonomy requires control over a wider range of action
than rule based systems can support. The subtle changes in flight
patterns, identification of new threats, self-directed changes in
mission profile and target selection all require autonomous assets
to go beyond pre-ordained instructions. Machine Learning and AI techniques
offer a viable way for autonomous systems to learn and evolve behaviors
that go beyond their programming. Semantic information passing from
sensors, via a sensor bus, to a collection of decision making controllers
makes provides for plug and play replacements of individual controllers.
An artificial neural network such as an RNN can ensemble and combine
inputs from multiple controllers to create a single, coherent control
signal. In taking this approach, while some of the individual controllers
may be rules-based, the RNN really evolves into the autonomous intelligence
that can consider a variety of concerns and factors via control system
inputs, and decide on the most optimum action. We propose delinking
control networks from the ensembler RNN so that individual control
RNNs may be evolved and trained to execute differing mission profiles
optimally, and these ``personalities'' may be easily uploaded into
the autonomous asset with no hardware changes necessary. One of the
challenges in taking this advanced approach may be the inability to
guarantee what exactly a learning, evolving autonomous system might
do. The action filter architecture proposed in this paper, which provides
a hard ``not to exceed'' boundary to range of action, delivers an
out-of-band method to audit and edit autonomous behavior, while still
keeping it within parameters of acceptability.


\begin{thebibliography}{10}
\bibitem{key-2}Vinson (2013). X-47B Makes First Arrested Landing
at Sea, Navy.mil.

\bibitem{key-3}Casteli et. al. (2016). Autonomous navigation for
low-altitude UAVs in urban areas, arxiv.org.

\bibitem{key-4}Ansari \& Alam (2011). Hybrid Genetic Algorithm fuzzy
rule based guidance and control for launch vehicle, Intelligent Systems
Design and Applications (ISDA) Conference.

\bibitem{key-5}Wang et. al. (2016). Learning to Reinforcement Learn,
arxiv.org.

\bibitem{key-6}HistoryNet (2006). Lawrence Sperry: Autopilot Inventor
and Aviation Innovator, HistoryNet.

\bibitem{key-7}Chicago Tribune (1947). Reveal 'Robot' C-54 Zig-Zagged
Way To England, Chicago Tribune Sept. 24, 1947.

\bibitem{key-8}Welch \& Bishop (2001). An Introduction to the Kalman
Filter, SIGGRAPH 2001.

\bibitem{key-9}Bin \& Justice (2009). The Design of an Unmanned Aerial
Vehicle Based on the ArduPilot, Indian Journal of Science \& Technology
April 2009.

\bibitem{key-10}Ferrer (2016). The blockchain: a new framework for
robotic swarm systems, arxiv.org.

\bibitem{key-11}Hexmoor et. al. (2005). Swarm Control in Unmanned
Aerial Vehicles, ICAI 2005.

\bibitem{key-12}Federal Aviation Administration. Flight Operation
Branch, Category I/II/III ILS information, www.faa.gov. 

\bibitem{key-13}Kennedy \& Eberhart (1995). Particle Swarm Optimization,
Proceedings of IEEE International Conference on Neural Networks. IV.
pp. 1942\textendash 1948.

\bibitem{key-14}Karaboga (2005). An Idea Based On Honey Bee Swarm
for Numerical Optimization, Technical Report-TR06, Erciyes University,
Engineering Faculty, Computer Engineering Department 2005.

\bibitem{key-15} Pan \& Yang (2009). A Survey on Transfer Learning,
IEEE Transactions on Knowledge and Data Engineering. 

\bibitem{key-16} Luotsinen (2004). Autonomous Environmental Mapping
In Multiagent UAV Systems, Masters Thesis, University of Central Florida. 
\end{thebibliography}
\end{document}